# Development of the InBan_CIDO Ontology by Reusing the Concepts along with Detecting Overlapping Information

*Archana Patel[1], and Narayan C. Debnath[2]

[1,2]Department of Software Engineering, School of Computing and Information Technology, Eastern International University, Vietnam

[1*]Email id: archanamca92@gmail.com
[2]Email id: narayan.debnath@eiu.edu.vn

**Abstract.** The covid-19 pandemic is a global emergency that badly impacted the economies of various countries. Covid-19 hit India when the growth rate of the country was at the lowest in the last 10 years. To semantically analyze the impact of this pandemic on the economy, it is curial to have an ontology. CIDO ontology is a well-standardized ontology that is specially designed to assess the impact of coronavirus disease and utilize its results for future decision forecasting for the government, industry experts, and professionals in the field of various domains like research, medical advancement, technical innovative adoptions, and so on. However, this ontology does not analyze the impact of the Covid-19 pandemic on the Indian banking sector. On the other side, Covid19-IBO ontology has been developed to analyze the impact of the Covid-19 pandemic on the Indian banking sector but this ontology doesn't reflect complete information of Covid-19 data. Resultantly, users cannot get all the relevant information about the Covid-19 and its impact on the Indian economy. This article aims to extend the CIDO ontology to show the impact of Covid-19 on the Indian economy sector by reusing the concepts from other data sources. We also provide a simplified schema matching approach that detects the overlapping information among the ontologies. The experimental analysis proves that the proposed approach has reasonable results.

## 1 Introduction

The new business models and the different government regulatory interventions during the COVID-19 pandemic have led to various behavioral and structural changes in peoples' lives across the globe. These changes cut across the width of the financial landscape thereby posing challenges to the key functions of the banking sector as well. In addition to other challenges, the major technological challenge is the inability to access systems and data because of various constraints. The existing Artificial Intelligence technologies should be leveraged to their full potential to enable data integration, analysis, and sharing for remote operations. Especially for financial institutions like the banking sector, artificial intelligence could offer better data integration and sharing through semantic knowledge representation approaches. Many reports containing plenty of data stating the effects of this pandemic on the banking sector are available in the public domain as listed below:

- https://dea.gov.in/
- https://finmin.nic.in/
- https://www.indiabudget.gov.in/economicsurvey/
- https://www.who.int/emergencies/diseases/
- https://www.mygov.in/covid-19/

These websites offer a static representation of Covid data which is largely unstructured in nature (text, audio, video, image, newspaper, blogs, etc.) creating a major problem for the users to analyze, query, and visualize the data. The data integration task gets highly simplified by the incorporation of knowledge organization systems (taxonomy, vocabulary, ontology) as background knowledge. Storing the knowledge using the semantic data models enhances the inference power also. Ontology is a data model that is very useful to enable knowledge transfer and interoperability between heterogeneous entities due to the following reasons: (i) they simplify the knowledge sharing among the entities of the system; (ii) it is easier to reuse domain knowledge and (iii) they provide a convenient way to manage and manipulate domain entities and their interrelationships. An ontology consists of a set of axioms (a statement is taken to be true, to act as a premise for further reasoning) that impose constraints or restrictions on sets of classes and relationships allowed among the classes. These axioms offer semantics because by using these axioms machines can extract additional or hidden information based on data explicitly provided [1]. Web ontology language (OWL) is designed to encode rich knowledge about the entities and their relationships in a machine-understandable manner. OWL is based on the Description Logic (DL) which is a decidable subset (fragment) of First-Order Logic (FOL) and it has a model-theoretic semantics.

Various ontologies have been developed in OWL language to semantically analyze the Covid-19 data. CIDO ontology is a well-standardized ontology and specially designed for coronavirus disease [2]. However, this ontology does not analyze the impact of the Covid-19 pandemic on Indian banking sectors. On the other side, Covid19-IBO ontology has been developed to analyze the impact of the Covid-19 pandemic on the Indian banking sector but this ontology doesn't reflect the complete information of Covid-19 [3]. The contributions of this article are as follows:

- To extend the CIDO ontology to encode the rich knowledge about the impact of Covid-19 on the Indian banking sector
- To detect the overlapping information by designing a simplified schema matching approach

The rest of the part of this article is organized as follows: Section 2 illustrates the available covid-19 ontologies. Section 3 focuses on the development of the proposed InBan_CIDO ontology. Section 4 shows the proposed schema matching approach, and the last section concludes this article.

## 2 A Glance on Available Covid-19 Ontologies

Ontology provides a way to encode human intelligence in a machine-understandable manner. For this reason, ontologies are used in every domain specifically in the emergency domain. As Covid-19 pandemic is a global emergency and various countries are badly impacted by it. India is one of the countries that is badly impacted by 2nd wave of this pandemic. Various ontologies are offered to semantically analyze the covid-19 data. These ontologies are listed below:

- Infectious Disease Ontology (called IDO) is an interoperable ontology that contains the domain information about infectious diseases. The entities of IDO ontology are related to the clinical and biomedical aspects of the disease [4]. IDO ontology provides a strong foundation to the other ontologies therefore this ontology is extended by various ontologies namely VIDO, CIDO, IDO-COVID-19, etc. The extended ontologies only import those concepts or entities that are required as per domain.

  The Coronavirus Infectious Disease Ontology (CIDO) is a community-based ontology that imports Covid-19 pandemic-related concepts from the IDO ontology [2]. CIDO is a more specific standard ontology as compared to IDO and encodes knowledge about the coronavirus disease as well as provides integration, sharing, and analysis of the information. The latest version of CIDO is released in May 2021.
- Dutta and DeBellis [5] have published the ontology as a data model namely COviD-19 ontology for the case and patient information (called CODO) on the web as a knowledge graph that provides the information about the Covid-19 pandemic. The primary focus of the CODO ontology is to describe the Covid-19 cases and Covid-19 patient data.
- Mishra et al. [3] have developed an ontology called Covid19-IBO ontology that semantically analyses the impact of Covid-19 on the performance of the Indian banking sector. This is only one ontology that reflects the Covid-19 information as well as its impact on the Indian economy.

**Table 1.** Metrics value of the available ontologies

| Metrics | Ontologies | | | | | |
|---|---|---|---|---|---|---|
| | IDO | VIDO | CIDO | IDO-Covid-19 | CODO | Covid19-IBO |
| Axioms | 4103 | 4647 | 104336 | 5018 | 2047 | 313 |
| Logical axioms count | 833 | 956 | 17367 | 1032 | 917 | 137 |
| Class count | 362 | 429 | 7564 | 486 | 91 | 105 |
| Object property count | 43 | 43 | 409 | 43 | 73 | 28 |
| Data property count | 0 | 0 | 18 | 0 | 50 | 43 |
| Attribute Richness | 0.0 | 0.0 | 0.056 | 0.0 | 0.549 | 0.409 |
| Inheritance Richness | 1.237 | 1.258 | 1.547 | 1.242 | 1.010 | 0.895 |
| Relation Richness | 0.384 | 0.360 | 0.873 | 0.340 | 0.471 | 0.229 |

Table 1 shows the value of the metrics of the available Covid-19 ontologies. CIDO ontology has the highest number of classes as compared to other ontologies. These classes describe the concept of the domain. Properties (data and object) increase the richness of the ontology. The axioms-imposed restriction on the entities of the ontology and provide an ability to semantically infer the information of the imposed queries from the ontology.

As of now, various methodologies for the development of ontology are proposed [6]. The four most famous ontology methodologies namely TOVE, Enterprise Model Approach, METHONTOLOGY, and KBSI IDEF5 are presented in figure 1. It is quite clear that developing ontologies is not focused on understanding the engineering process but it is a matter of craft skill. The selection of the methodologies heavily depends on the application and requirement of the ontology. Therefore, there is no perfect methodology available that can be used universally.

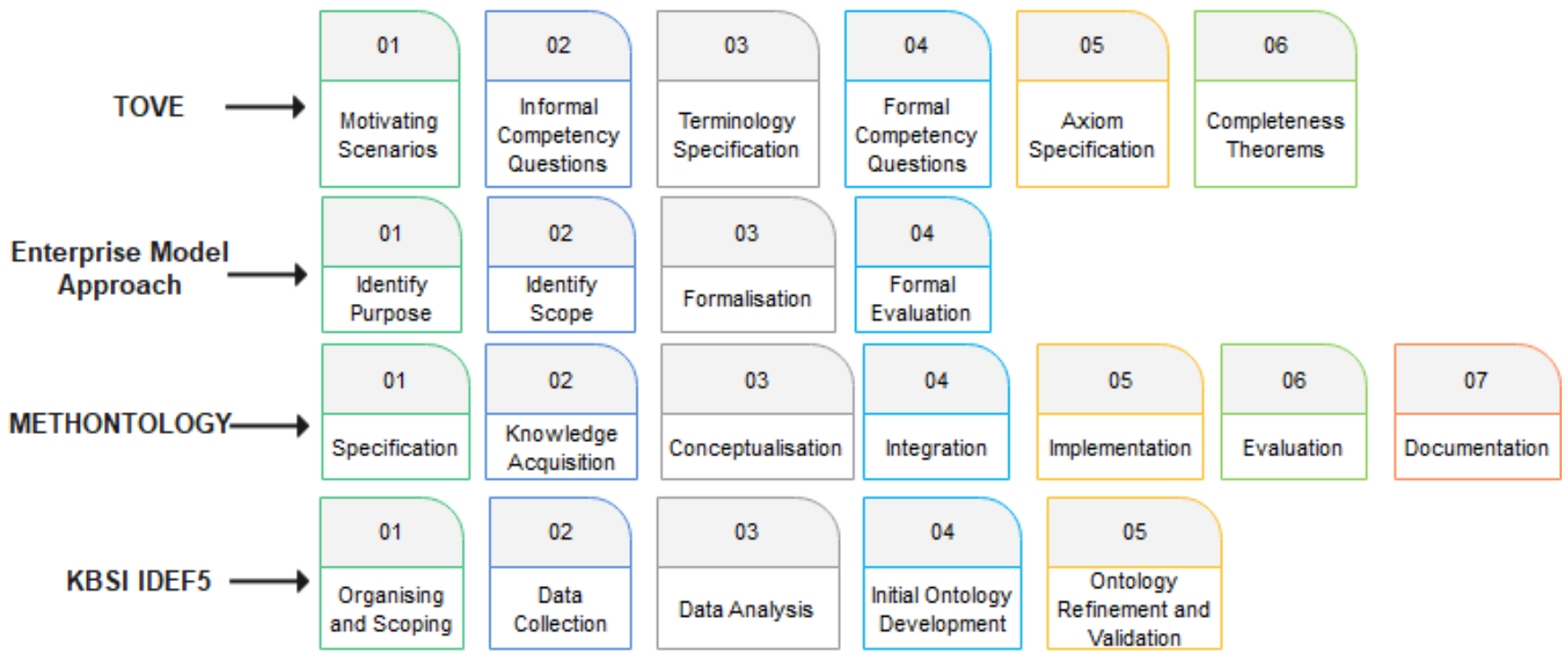

**Fig. 1.** Available methodologies and their Steps

There are two ways to create an entity inside the ontology, (a) to reuse concepts from the available ontologies (b) to create all concepts separately (without reusing concepts). It is good practice to reuse the existing concepts that offer a common understanding of the domain. With the help of IRI, we can reuse the concepts. IRI is internationalized resource identifier which avoids multiple interpretations of entities of ontology.

After studying the literature, we state that available ontologies contain detailed information about the Covid-19 disease. However, they do not have complete (or partial) information about the impact of Covid-19 on the Indian banking sector (the sector which plays a vigorous role in the growth of the Indian economy) along with Covid-19 data. The different Covid-19 ontologies create the heterogeneity problem that needs to resolve in order to achieve interoperability among the available ontologies.

## 3 InBan_CIDO Ontology

Ontology is a semantic model that represents the reality of a domain in a machine-understandable manner. The basic building block of an ontology is classes, relationships, axioms, and instances [7]. Nowadays, ontologies are used everywhere because of their ability to infer semantic information of the imposed queries. As literature shows available ontologies are not capable to analyse the impact of Covid-19 on the Indian economy along with Covid-19 detailed information. Therefore, we extend the CIDO ontology to fill this gap. On behalf of the above-listed methodologies, we have selected five phases for the extension of the CIDO ontology (called InBan_CIDO) that will offer complete information about the impact of Covid-19 on the performance of the Indian banking sector along with information on the Covid-19 pandemic. These five phases are Scope determination, Extraction of the concept, Organization of the concept, Encoding, and Evaluation. The detailed description of these phases is stated below:

- *Scope Determination:* The objective of this phase is to determine the scope of the ontology. We use competency questions to fix the scope and boundary of the proposed ontology. Some selected questions are mentioned below that are framed after the discussion with the expert of the domain.
  a) How to the central bank will tackle the situation that arises due to non-collection of debt recovery in the moratorium period during the Covid-19 lock down.
  b) What will be the impact on the balance sheet of banks when the NPA number will be added for the period of Covid-19.
  c) What is the cumulative impact on the banking industry due to the loss of other industries like aviation, tourism, marketing, etc?
  d) How risk assessment and planning should be done for the upcoming Covid wave (if any).
  e) How effectively the risk assessment and mitigation mechanism worked during the first and second waves.

- *Concept Extraction:* This phase aims to extract the concepts or entities from the different sources as per the specified domain. These concepts can be marked as classes, properties (data and object properties), and instances in the further phases. We use the following sources for the development of the InBan_CIDO ontology.
  a) Research articles from various data sources and indexed by Scopus, SCI, etc
  b) Ontology repositories like OBO library, Bio portal, etc
  c) Ontologies like Covid19-IBO
  d) Databases provided by WHO and Indian government
  e) Interview with the experts of the domain

  We have extensively gone through these data sources and extracted all the entities that are required for the extension of CIDO ontology to fulfill the proposed scope. All the extracted entities are stored in the excel sheet for further analysis.

- *Concept Organization:* This phase aims to organize the extracted concepts in a hierarchical manner. Firstly, we classified the extracted concepts, as classes, properties, and instances based on their characteristics. All the identified concepts are organized in a hierarchal manner (parent-child relationship). For example, a class *private bank* and a class *government bank* should become the subclasses of class *Bank*. We import some concepts inside InBan_CIDO ontology from the other ontologies like Covid19-IBO. Some imported concepts are mentioned below:

**Reused Classes**

Current_Challenging_conditions_of_banking_industry, NPA, Loan, Bank, InfectedFamilyMember, Scheduled_Banks, Doubtfull, ETB, Cooperative_Banks, Impact, Commercial_Banks, Financial, Employee, Bankers, Digital_optimization, Loan_repayment, NRE, Detect_probable_defaults_in_early_phase, New_Assets_Quality_Review, IndividualCurrentAccount, Robust_digital_channels, OnHuman, BankingRetailCenters, Contactless_banking_options, High_credit_risk, Private_Sector, Deposit, Policies

**Reused data properties**

ContainedIn, EmployeeID, hasBankingRelationship, has_status, has_temp_of_human, number_of_account, number_of_credit_card, number_of_loans_reported, sanctioned_strength, working_strength, has_date

**Reused object properties**

Return, negative_return, positive_return, hasStatistics, has_cause, has_close, has_gender, has_nationality, has_open, type_of_relationship, via_account, via_card, via_loan, via_insurance, city_wise_statistics

- *Encoding:* We encode the InBan_CIDO ontology by using protégé 5.5.0 tool [8]. Protégé tool is free available on the web and it has a very interactive interface. User can encode the ontology in protégé without having any technical knowledge about any programming language. InBan_CIDO ontology has two types of classes: old classes and new classes. Old classes are those classes that already available in CIDO ontology (CIDO is the base ontology that we have extended). New classes are categories into two groups:

  a) Classes are imported inside InBan_CIDO from other available ontologies (source ontologies) by using the IRI of that ontology
  b) New classes are added as per need just by creating classes under the thing class

  *The process of Importing Classes inside InBan_CIDO Ontology:* For the reusability purpose, we import some classes in the InBan_CIDO ontology from the Covid19-IBO ontology. The process to import the classes inside destination ontology is required the IRI of that source ontology where these classes are defined. After getting the IRI of the ontology, we open the protégé tool and create the name of the class under the thing class (which is a default class) or any other classes as per need then go to *the new entity option* (figure 2.a) and click on *specified IRI* (figure 2.b) and write the IRI of the source ontology where that concept defined and then click on the ok option. Now, the IRI of the class will be changed (figure 2.c). For example, figure 2 shows the process to import the class *Person* in InBan_CIDO ontology from the FOAF ontology which is an upper ontology. The class *person* of Covid19-IBO ontology is also imported from FOAF ontology [9].

To define the new classes inside InBan_CIDO ontology, we simply click on Active Ontology IRI (fig 2.b) instead of the specified IRI which we did in the case of importing the concepts. Figure 3 shows the screenshot of the InBan_CIDO ontology that has been taken from the plugins OWLViz of protégé tool. The latest version of InBan_CIDO ontology is available on the bio-portal for public use.

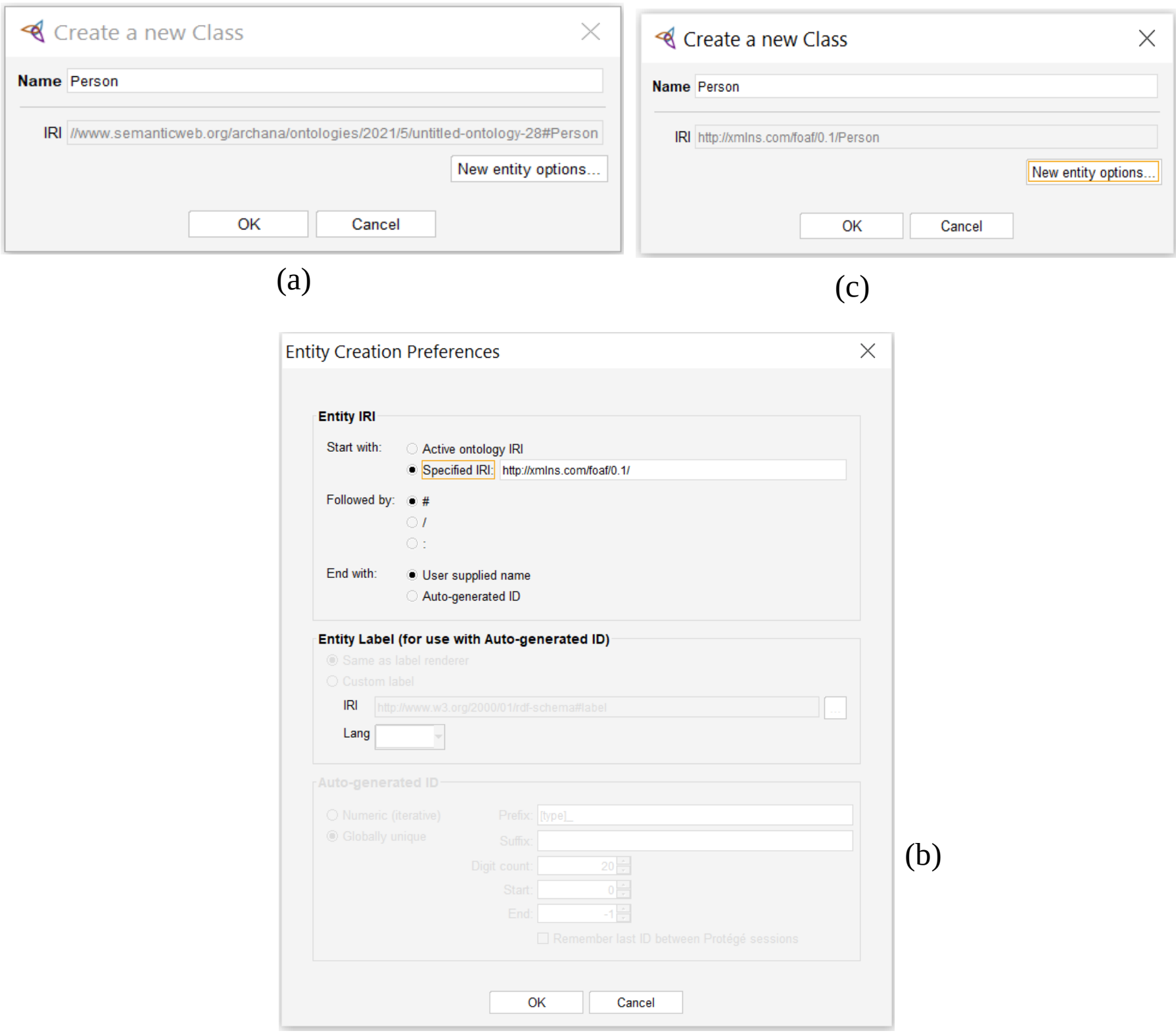

**Fig. 2.** The process to import the concepts in InBan_CIDO Ontology

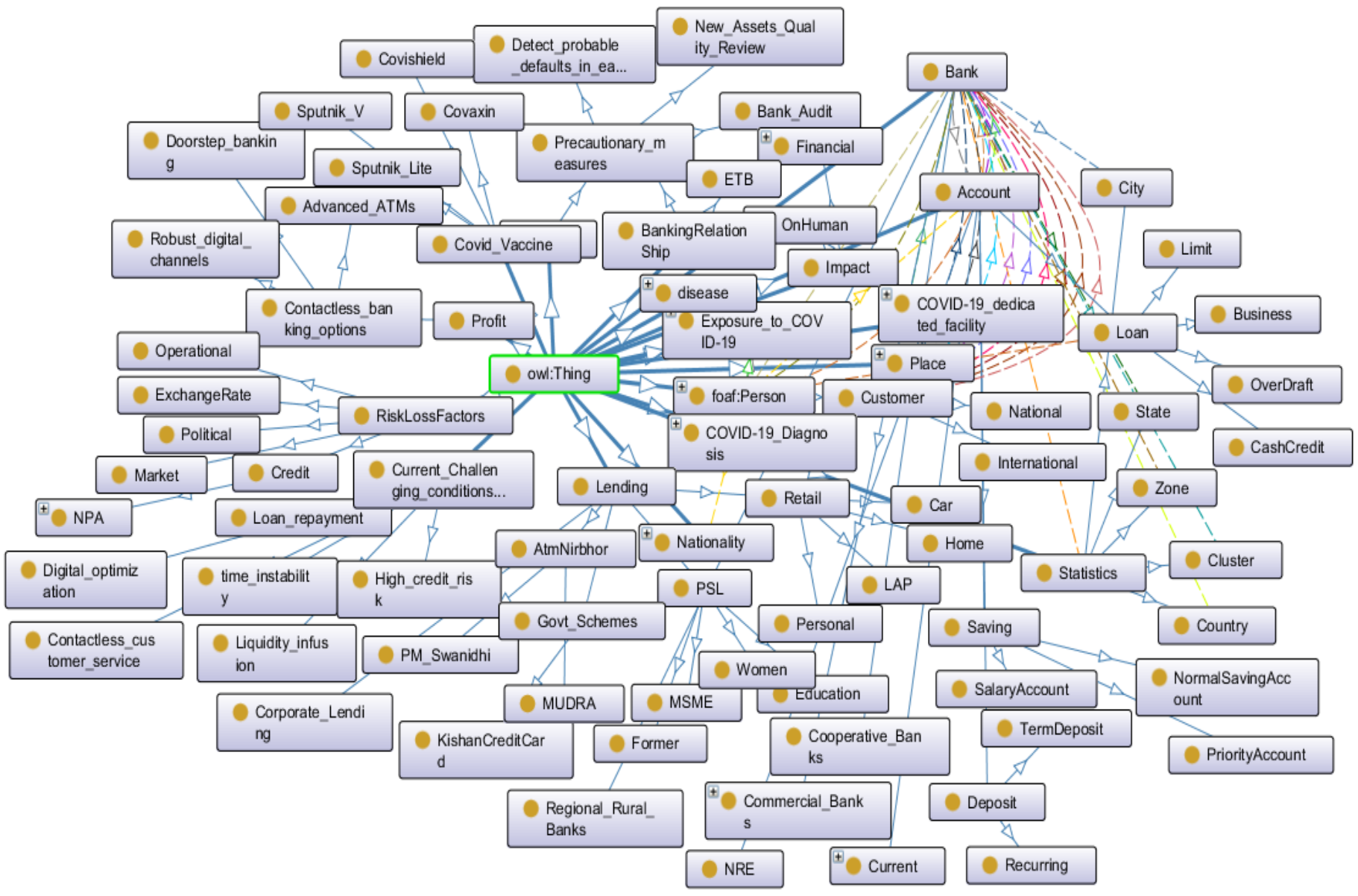

**Fig. 3.** InBan_CIDO Ontology

- *Evaluation:* Evaluation of an ontology determines the quality, completeness, and correctness of an ontology according to the proposed scope. We have used the OOPs tool [10] to know the available anomalies (pitfall) inside the InBan_CIDO ontology. OOPs detects the pitfall in three categories namely Critical, Minor and Important. The critical pitfall is very serious because it damages the quality of the ontology. Figure 4 shows that there is no critical pitfall in the InBan_CIDO ontology and all the minor and important pitfalls are removed by extending the ontology.

| | |
|---|---|
| Results for P04: Creating unconnected ontology elements. | 6 cases \| Minor |
| Results for P07: Merging different concepts in the same class. | 1 case \| Minor |
| Results for P08: Missing annotations. | 236 cases \| Minor |
| Results for P10: Missing disjointness. | ontology* \| Important |
| Results for P11: Missing domain or range in properties. | 1 case \| Important |
| Results for P13: Inverse relationships not explicitly declared. | 35 cases \| Minor |
| Results for P22: Using different naming conventions in the ontology. | ontology* \| Minor |
| Results for P30: Equivalent classes not explicitly declared. | 2 cases \| Important |
| Results for P41: No license declared. | ontology* \| Important |

**Fig. 4.** OOPs Results of InBan_CIDO ontology

## 4 A Simplified Schema Matching Approach for Covid-19 Ontologies

The developed InBan_CIDO ontology contains information about the Covid-19 disease as well as information of the impact of Covid-19 on the performance of the Indian banking sector which is imported from Covid19-IBO ontology. The Covid19-IBO ontology has also contained information about the Covid-19 disease as per need. Therefore, it is required to investigate the overlapping information from the existing Covid-19 ontologies with respect to Covid19-IBO ontology. Matching systems use the matching algorithm and detect the relationships between the entities of the ontologies [11].

We propose a Schema Matching Approach (SMA-Covid19) to find out the overlapping information among the developed Covid-19 ontologies. The first step of the SMA-Covid19 algorithm is to select two ontologies ($O_S$ and $O_T$) from the ontology repository and then extract all the labels of the concepts in both ontologies with the help of Id and IRI. The labels of the concepts of source and target ontologies are stored in a n, m dimensional array (a[n] and b[m]) separately where, n=number of classes in $O_S$ and m=number of classes in $O_T$. The SMA-Covid19 algorithm picks one label of $O_S$ and then matches it with all the labels of $O_T$. The matching between the labels is performed according to the Levenshtein and synonym matchers. If two labels are matched based on synonyms, then a 0.9 similarity value will be assigned to them. The matching result (similarity value between the labels) is stored in the n×m matrix (Avg[n][m]). All the pairs whose similarity value is greater than $\alpha$ (where $\alpha$ is a threshold for the similarity value) are considered to be correspondences.

**Pseudocode for SMA-Covid19**

- Select $O_S$ and $O_T$ from the ontology's repository
- Extract all labels of both $O_S$ and $O_T$ ontologies

```
for (i=1, i ≤ n, i++)   // n∈ number of classes in O_S
    fetch Id of the i^th concept
        if (Id has label)
            fetch and store label in a[i]
        else
            fetch IRI of the i^th concept
            and find label by splitting IRI
            store label in a[i]
```

```
for (j=1, j ≤ m, i++)   // m∈ number of classes in O_T
    fetch Id of the j^th concept
        if (Id has label)
            fetch and store label in b[j]
        else
            fetch IRI of the j^th concept
            and find label by splitting IRI
            store label in b[j]
```

- Matching

```
for (i=1, i<n, i++)
    for (j=1, j<m, j++)
        Sim1 ← Run synonym matcher over a[i] and b[j]
        Sim2 ← Run Levenshtein matcher over a[i] and b[j]
        Avg[i][j] ← Average of Sim1 and Sim2
Display all the pairs where Avg[i][j] ≥ α
```

*Experimental configuration and Analysis:* We have run SMA-Covid19 in windows server 2012 R2 standard with an Intel Xeon, 2.20 GHz (40 cores) CPU, and 128 GB RAM. The proposed approach is implemented in the Python programming language. During experiments, we have set parameter ∝=0.8. We used libraries like

Numpy (NumPy is a Python library, supporting large, multi-dimensional arrays and matrices, along with a large number of high-level mathematical functions for these arrays) and pandas (Open source, BSD-licensed library providing high-performance, easy-to-use data structures and data analysis tools for the Python programming language). We have imported the Python library for string matching and fetched a list of words from the nltk library (from nltk.corpus import stopwords).

Figure 5 shows the matching results of available Covid-19 ontologies namely CODO, VIDO, CIDO, IDO, IDO-COVID-19 with respect to Covid19-IBO ontology in terms of performance parameters namely precision, recall, and F-measure. The precision parameter explains the correctness of the algorithm whereas the Recall parameter measures the completeness of the algorithm. The parameter F-measure is the harmonic mean of precision and recall [12]. The obtained results of these parameters show that the proposed matching algorithm has reasonable performance.

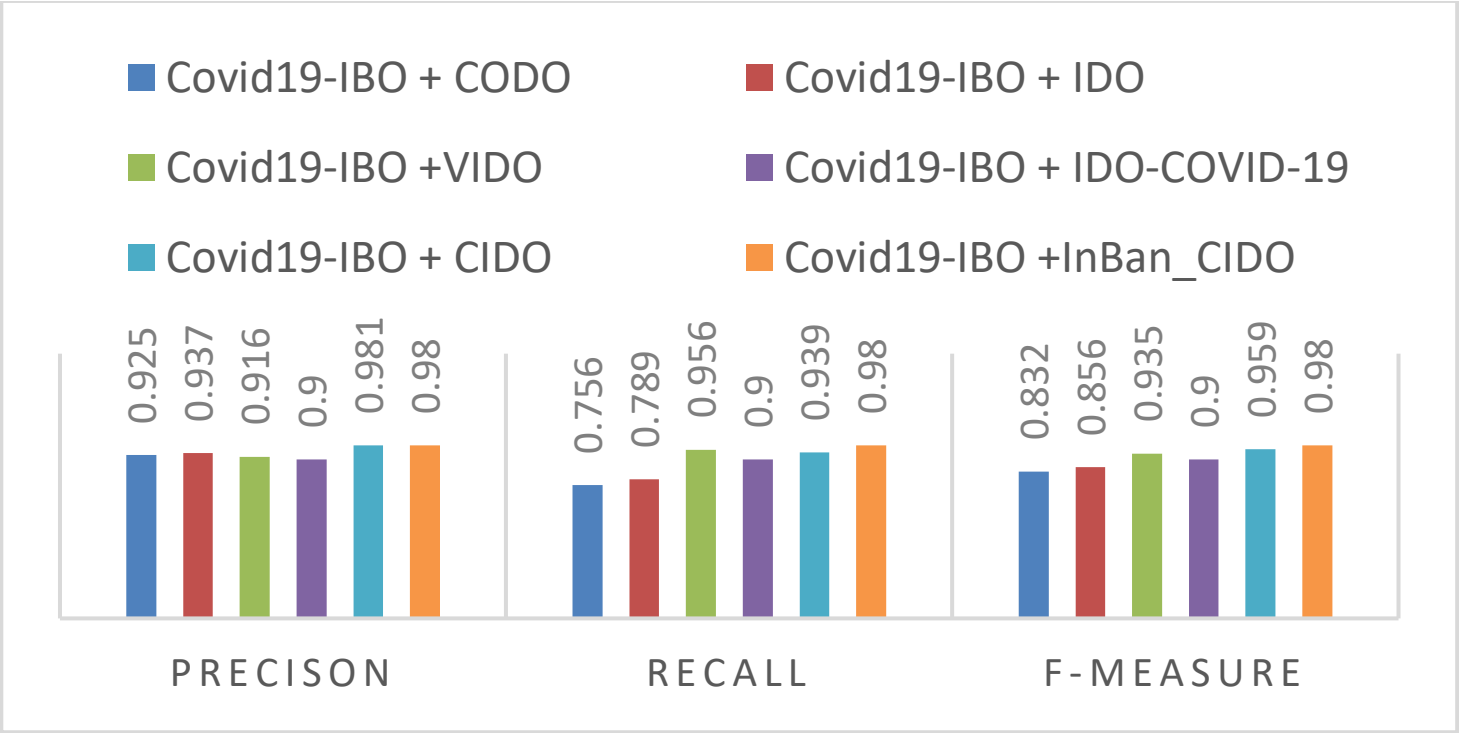

**Fig. 5.** Performance Result of SMA-Covid19 Approach

## Conclusion

We have extended the CIDO ontology by reusing the concepts from other data sources. The developed InBan_CIDO ontology offers accurate and precise knowledge about the impact of Covid-19 on the Indian economy as well as detailed information about the Covid data. To detect the overlapping information, we have provided the SMA-Covid19 algorithm approach that matches the schema of Covid-19 ontologies. The experimental analysis shows that the proposed approach has reasonable results in terms of precision, recall, and F-measure.